\documentclass[conference]{IEEEtran}
\IEEEoverridecommandlockouts
\usepackage{cite}
\usepackage{amsmath,amssymb,amsfonts}
\usepackage{algorithmic}
\usepackage{graphicx}
\usepackage{textcomp}
\usepackage{xcolor}
 
\usepackage{multirow}
\usepackage{booktabs} 
\usepackage{url}
 
\def\BibTeX{{\rm B\kern-.05em{\sc i\kern-.025em b}\kern-.08em
    T\kern-.1667em\lower.7ex\hbox{E}\kern-.125emX}}
\begin{document}

\title{Defense Against Adversarial Attacks with Saak Transform}


\author{%
\IEEEauthorblockN{Sibo Song\IEEEauthorrefmark{1}, Yueru Chen\IEEEauthorrefmark{2}, Ngai-Man Cheung\IEEEauthorrefmark{1}, C.-C. Jay Kuo\IEEEauthorrefmark{2}} \\
\IEEEauthorblockA{\IEEEauthorrefmark{1} \textit{Information Systems Technology and Design}, \textit{Singapore University of Technology and Design}, Singapore, Singapore}
\IEEEauthorblockA{\IEEEauthorrefmark{2} \textit{Ming Hsieh Department of Electrical Engineering}, 
\textit{University of Southern California}, Los Angeles, California, USA}
}

\maketitle

\begin{abstract}
Deep neural networks (DNNs) are known to be vulnerable to adversarial perturbations, which imposes a serious threat to DNN-based decision systems. In this paper, we propose to apply the lossy Saak transform to adversarially perturbed images as a preprocessing tool to defend against adversarial attacks. Saak transform is a recently-proposed state-of-the-art for computing the spatial-spectral representations of input images. Empirically, we observe that outputs of the Saak transform are very discriminative in differentiating adversarial examples from clean ones. Therefore, we propose a Saak transform based preprocessing method with three steps: 1) transforming an input image to a joint spatial-spectral representation via the forward Saak transform, 2) apply filtering to its high-frequency components, and, 3) reconstructing the image via the inverse Saak transform. The processed image is found to be robust against adversarial perturbations.  We conduct extensive experiments to investigate various settings of the Saak transform and filtering functions. Without harming the decision performance on clean images, our method outperforms state-of-the-art adversarial defense methods by a substantial margin on both the CIFAR-10 and ImageNet datasets. Importantly, our results suggest that adversarial perturbations can be effectively and efficiently defended using state-of-the-art frequency analysis.
\end{abstract}


\section{Introduction}\label{sec:intro}

Recent advances in deep learning have made unprecedented success in many
real-world computer vision problems such as face recognition, autonomous
driving, person re-identification, etc.  \cite{schroff2015facenet,
xu2016end,li2014deepreid}. However, it was first pointed by Szegedy {\em
et al.} \cite{szegedy2013intriguing} that deep neural networks (DNN) can
be easily fooled by adding carefully-crafted adversarial perturbations
to input images. These adversarial examples can trick deep learning
systems into erroneous predictions with high confidence.  It was
further shown in \cite{kurakin2016adversarial} that these examples exist
in the physical world. Even worse, adversarial attacks are often
transferable \cite{kurakin2016badversarial,liu2016delving}; namely, one
can generate adversarial attacks without knowing the parameters of a
target model. 

These observations have triggered broad interests in adversarial defense
research to improve the robustness of DNN-based decision systems.
Currently, defenses against adversarial attacks can be categorized into
two major types. One is to mask the gradients of the target neural
networks by modifying them through adding layers or changing
loss/activation functions, e.g., \cite{papernot2016distillation,
ross2017improving,lyu2015unified}. The other is to remove adversarial
perturbations by applying transformations to input data
\cite{guo2017countering, xu2017feature, liu2018feature,das2017keeping}.
Since these transformations are non-differentiable, it is difficult for
adversaries to attack through gradient-based methods. 

\begin{figure*}
\centering
\includegraphics[width=1.8\columnwidth]{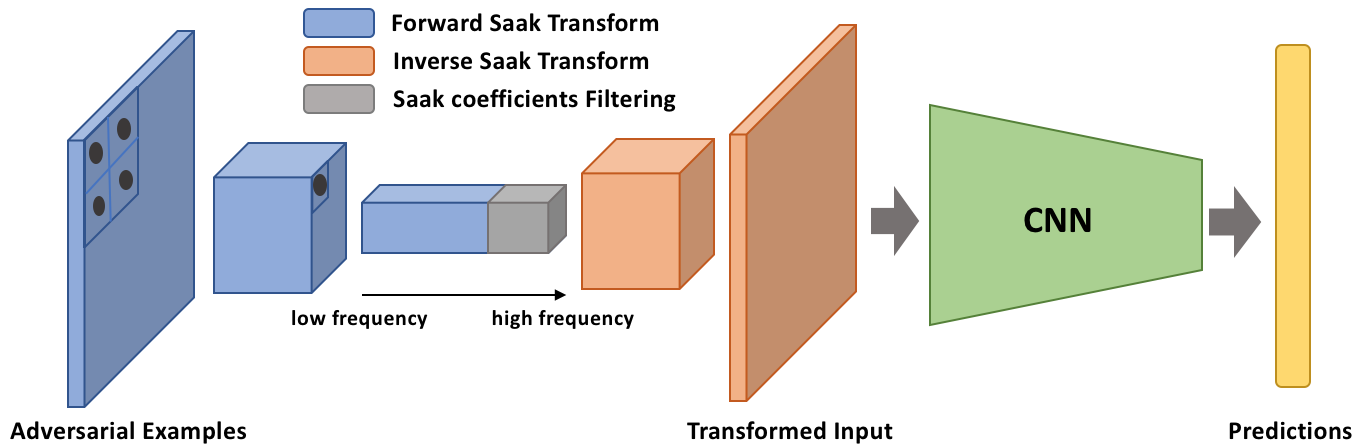}
\caption{Demonstration of our proposed method to filter out the
adversarial perturbations with the multi-stage Saak transform.}
\label{fig:diagram}
\end{figure*}

In this work, we focus on adverserial attacks to the convolutional
neural network (CNN) and propose a defense method that maps input images
into a joint spatial-spectral representation with the forward Saak
Transform \cite{kuo2018data}, purifies their spatial-spectral
representations by filtering out high-frequency components and, then,
reconstructs the images. As illustrated in Fig.  \ref{fig:diagram}, the
proposed mechanism is applied to images as a preprocessing tool before
they goes through the CNN.  We explore three filtering strategies and
apply them to transformed representations to effectively remove
adversarial perturbations.  The rationale is that, as adversarial
perturbations are usually undetectable by the human vision system (HVS),
reducing high-frequency components should contribute to adversarial
noise removal without hurting the decision accuracy of clean data much
since it preserves components that are important to the HVS in restored
images. 
We propose to use the Saak Transform \cite{kuo2018data} to perform the frequency analysis. 
As the state-of-the-art for computing spatial-spectral representation, our empirical results demonstrate that Saak coefficients of high spectral dimensions are discriminative for adversarial and clean examples. 
Our algorithm 
is efficient since it demands neither adversarial
training with any label information nor modification of neural networks. 

\section{Related Work}\label{sec:review}

Adversarial attacks have been extensively studied since the very
earliest attempt \cite{szegedy2013intriguing}, which generated L-BFGS
adversarial examples by solving a box-constrained optimization problem
to fool the neural networks.  Goodfellow {\em et al.}
\cite{goodfellow2014explaining} developed an efficient fast gradient
sign method (FGSM) to add adversarial perturbations by computing the
gradients of the cost function w.r.t the input.  Along this direction,
Kurakin {\em et al.} \cite{kurakin2016adversarial} proposed a basic
iterative method (BIM) that iteratively computes the gradients and takes
a small step in the direction (instead of a large one like the FGSM).
Later, Dong {\em et al.} \cite{dongboosting} integrated a momentum term
to the BIM to stabilize update directions.  Papernot {\em et al.}
\cite{papernot2016limitations} generated an adversarial attack by
restricting the $L_0$-norm of perturbations.  DeepFool (DF)
\cite{moosavi2016deepfool} iteratively calculates perturbations to take
adversarial images to the decision boundary which is linearized in the
high-dimensional space. It was further extended to fool a network with a
single universal perturbation \cite{moosavi2017universal}. Carlini and
Wagner \cite{carlini2017towards} proposed three variants of $CW$
adversarial attacks under $L_0$, $L_2$ and $L_{\inf}$ distance
constraints.  Chen {\em et al.} \cite{chen2017ead} generated a strong
$L_1$ attack by adding elastic-net regularization to combine $L_1$ and
$L_2$ penalty functions.  Unlike above-mentioned methods, Xiao {\em et
al.} \cite{xiao2018spatially} proposed a novel method that crafts
adversarial examples by applying spatial and locally-smooth
transformation instead of focusing on pixel-level change.  Su {\em et
al.} \cite{su2017one} presented a way to attack neural networks by
changing only one pixel from each image through a differential evolution
algorithm. 

Recently, a few defense techniques have been proposed in detecting and
defending against adversarial attacks. Papernot {\em et al.}
\cite{papernot2016distillation} proposed a defensive distillation method
that uses soft labels from a teacher network to train a student model.
Gu and Rigazio \cite{gu2014towards} applied stacked denoising
auto-encoders to reduce adversarial perturbations. Li and Li
\cite{li2016adversarial} used cascaded SVM classifiers to classify
adversarial examples. They also showed that $3\times3$ average filters
can mitigate adversarial effect. Recently, Tramer {\em et al.}
\cite{tramer2017ensemble} achieved good results by training networks
with adversarial images generated from different models.  Guo {\em et
al.} \cite{guo2017countering} applies input transformations such as
cropping, bit-depth reduction, JPEG compression and total variance
minimization to remove adversarial perturbations.  Similarly, Xu {\em et
al.} \cite{xu2017feature} proposed several strategies, including median
smoothing and bit-depth reduction, to destruct adversarial perturbations
spatially.  Dziugaite {\em et al.} \cite{dziugaite2016study}
investigated the effect of JPEG compression on adversarial images.
Based on that, Akhtar {\em et al.} \cite{akhtar2017defense} built a
perturbation detector using the Discrete Cosine Transform (DCT).  Very
recently, Liu {\em et al.} \cite{liu2018feature} designed an
adversarial-example-oriented table to replace the default quantization
table in JPEG compression to remove adversarial noise. 

The Saak transform \cite{kuo2018data} provides an efficient, scalable
and robust tool for image representation.  It is a representation not
derived by differentiation.  This makes gradient-based or
optimization-based attack difficult to apply.  The Saak transform has
several advantages over the DCT in removing adversarial perturbations.
First, Saak transform kernels are derived from the input data while the
ones used in DCT are data independent.  Saak kernels trained with clean
data from a specific dataset are more effective in removing perturbation
noise. Furthermore, the PCA is used in the Saak transform to remove
statistical dependency among pixels, which is optimal in theory. The DCT
is known to be a low-complexity approximation to the PCA in achieving
the desired whitening effect.  Finally, the multi-stage Saak transform
can preserve the prominent spatial-spectral information of the input
data, which contributes to robust classification of attacked images.  In
this work, we investigate Saak coefficients of each spectral component,
and observe that high frequency channels contribute more to adversarial
perturbations. We conduct experiments on the CIFAR-10 and the ImageNet
datasets and show that our approach outperforms all state-of-the-art
approaches. 

\section{Proposed Method}\label{sec:method}

\subsection{Problem Formulation}

Before introducing the proposed method, we formulate the problem of
adversarial attacks and defenses with respect to a given neural network.
A neural network is denoted by a mapping $\mathbf{y} = f(\mathbf{x})$,
where $\mathbf{x}\in\mathbb{R}^{W\times H\times D}$ is and input image
of size $W\times H\times D$ and $\mathbf{y}\in\mathbb{N}$ is the
predicted output vector. Given neural network model $f$, clean image
$\mathbf{x}$ and its ground-truth label $\hat{\mathbf{y}}$, crafting an
adversarial example denoted by $\mathbf{x'}$ can be described as a 
box-constraint optimization problem:
\small
\begin{equation}
\min _{ \mathbf{x}' }{ \left\|\mathbf{x}'-\mathbf{x}\right\| }_{p}, 
\quad \text{s.t.} \quad \mathbf{y} = \hat{\mathbf{y}}, \quad 
f(\mathbf{x}')\neq \hat{\mathbf{y}},
\end{equation}
\normalsize
where $\left\| \cdot \right\|$ is the $L_p$ norm.  Our goal is to find a
transformation function $\Psi(\cdot)$ to mitigate the adversarial effect
of $\mathbf{x}'$. In other words, we aim at obtaining a transformation
such that predictions of the transformed adversarial examples are as
close to ground-truth labels as possible. Ideally, $f(\Psi
(\mathbf{x}'))=f(\Psi (\mathbf{x}))=f(\mathbf{x})=\hat { \mathbf{y} }$.
In most settings of recent attacks, an adversary can have direct access
to the model $f(\mathbf{x})$and attack the model by taking advantage of
the gradients of the network w.r.t the input. For this reason, a desired
transformation, $\Psi(\cdot)$, should be non-differentiable.  This will
make attacks on the target model $f(\Psi (\cdot))$ more challenging even
if an attacker can access all parameters in $f(\cdot)$ and
$\Psi(\cdot)$. 

\subsection{Image Transformation via Saak Coefficients Filtering}\label{subsec:idea} 

\begin{figure*}[htbp]
\centering
\includegraphics[width=0.465\columnwidth]{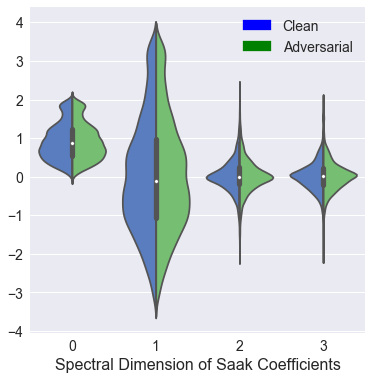} 
\includegraphics[width=0.492\columnwidth]{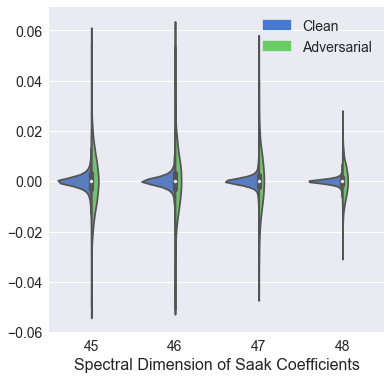} 
\includegraphics[width=0.9\columnwidth]{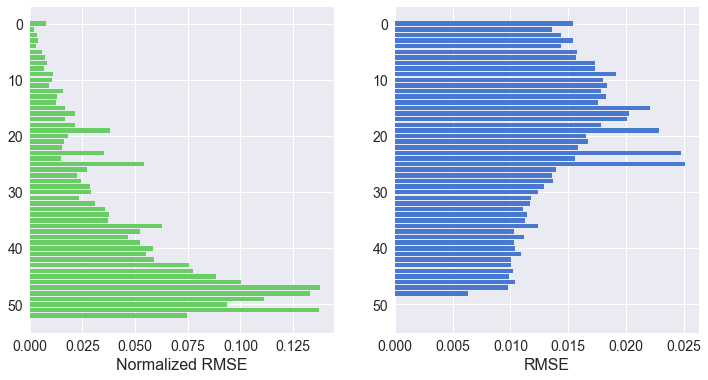}
\caption{From left to right: The distributions (histograms) of Saak coefficients of
clean images and FGSM adversarial examples in four representative low
and high spectral components are shown in (a) and (b), respectively.
In (a), distributions of Saak coefficients belonging to low spectral dimension 0 to 3 are shown.
In (b), distributions of Saak coefficients belonging to high spectral dimension 45 to 48 are shown.
The y-axis represents the values of Saak coefficients.
As shown in (b), for high spectral components, the Saak coefficients distributions are different between clean and adversarial examples.
In (c) and (d), the
normalized and the original RMSE (root-mean-squared-error) values
between clean and FGSM adversarial examples in different spectral
components are shown.
The y-axis in both figures represents the spectral dimension 
of Saak coefficients.
In (c), we can observe that clean and adversarial examples have different coefficient values in high spectral dimensions. 
(These results are obtained from two-stage Saak transform with $2\times2$ local cuboids.)
}\label{fig:fig_01}
\end{figure*}


{\bf Saak Transformation of an image.}
As shown in the left part of Fig. \ref{fig:diagram}, an image is
decomposed into blocks of $2\times2$ pixels (or local cuboids (LCs) of shape
$2\times2\times k$, where $k$ is the spectral dimension).  Then, the KLT
transform is conducted on the block (or cuboid) by merging four spatial
nodes into one parent node.  This process is recursively conducted stage
by stage in the forward multi-stage Saak transform. Note that, $2\times2$ is 
a typical size of LCs for illustration purpose, we can use arbitrary size in practice. The signed KLT coefficients in
each stage are called {\em Saak coefficients}. The Saak coefficients are discriminative
features of the input image. One can define a multi-stage inverse Saak
transform that converts cuboids of lower spatial resolutions and high
spectral resolutions to cuboids of higher spatial resolutions and lower
spectral resolutions and, eventually, reconstructs the approximate (or
original) image depending on whether the lossy (or lossy) Saak transform
is adopted.

{\bf Saak coefficients as discriminative
features.}
We compare the histogram of Saak coefficient values of clean (in blue)
and adversarial images (in green) in four representative low and high
spectral components in Figs. \ref{fig:fig_01} (a) and (b), respectively.
Following the experiment setup in \cite{liu2018feature}, 
the adversarial examples are generated using FGSM algorithms on the 
first 100 correctly classified samples in the CIFAR-10 test set.
We remark that 
similar observations can be found in other adversarial attacks.
We see that clean and adversarial images share similar distributions in 
lower spectral components.
However, for high spectral components, adversarial examples have much larger variances than
clean examples. These results indicate
that high-frequency Saak coefficients have discriminative power in
distinguishing adversarial and clean examples.  In addition, we
show the normalized and the original root-mean-square error of Saak
coefficients between clean images and adversarial examples in Figs.
\ref{fig:fig_01} (c) and (d), respectively. 
After normalization by the range of clean Saak coefficients, we
see clearly that the difference between clean and adversarially
perturbed images primarily lies in high-frequency Saak coefficients. 
Based on the above observation, we propose to use the multi-stage Saak
transform \cite{kuo2018data} as the transformation function
$\Psi(\cdot)$. As mentioned earlier, the Saak transform can offer a
joint spatial-spectral representation.  It maps a local 3D cuboid into a
1D rectified spectral vector via a one-stage Saak transform. Multiple
local 3D cuboids can be transformed in parallel, and the union of them
form a global 3D cuboid. The global cuboid consists of two spatial
dimensions and one spectral dimension.  The one-stage Saak transform
consists of two cascaded operations: 1) signal transform via PCA and 2)
sign-to-position format conversion.  It allows both lossless and lossy
transforms.  The distance between any two input vectors and that of
their corresponding output vectors is preserved at a certain degree.
Furthermore, the one-stage Saak transform can be extended to multi-stage
Saak transforms by cascading multiple one-stage Saak transforms so as to
provide a wide range of spatial-spectral representations and higher order
of statistics.

\subsection{Mathematical Formalization}

Mathematically, the Saak-based preprocessing technique can be written as
\small
\begin{equation}
\Psi (\mathbf{x'})=S^{-1} (\omega (S (\mathbf{x'}))),
\end{equation}
\normalsize
where $\mathbf{x}'$ is an adversarial example, $S(\cdot)$ and
$S^{-1}(\cdot)$ are the forward and inverse multi-stage Saak transforms,
respectively, and $\omega(\cdot)$ denote a filtering function to be
discussed later. Given Saak coefficients $\theta = S (\mathbf{x'})$ in
an intermediate stage, one can reconstruct an image using the inverse
Saak transform on filtered Saak coefficients denoted by $\omega (S
(\mathbf{x'}))$.  In other words, we convert an adversarial perturbation
removal problem into a Saak coefficients filtering problem. 
We attempt to formalize the idea in Sec. \ref{subsec:idea} below.

First, we have
\small
\begin{align}
\begin{split}
\left\| \Psi (\textbf{x}')-\Psi (\textbf{x}) \right\| _{ 2 } 
& =\left\| S^{-1} (\omega (\theta '))- S^{-1} (\omega (\theta )) \right\| _{ 2 } \\ 
& \approx \left\| \omega (\theta ')-\omega (\theta ) \right\|_{ 2 } \\ 
& =\left\| \omega (\theta _{ l }')-\omega (\theta _{ l }) \right\| _{ 2 } +
\left\| \omega (\theta _{ h }')-\omega (\theta _{ h }) \right\| _{ 2 },
\end{split}
\end{align}
\normalsize
where $\theta_h$ and $\theta_l$ denote the orthogonal subsets of high-
and low-frequency Saak coefficients. The approximation is based on the
semi-distance preserving property of the inverse Saak transform.  Under
the assumption that adversarial noise lies in high-frequency regions,
which is supported by the discussion in Sec. \ref{subsec:idea}, we
have 
\small
\begin{align}
\begin{split}
\left\| \Psi ({ \textbf{x} }')-\Psi ({ \textbf{x} }) \right\| _{ 2 } 
& =\left\| \theta _{ l }'-\theta _{ l } \right \| _{ 2 } + \left\| \omega 
(\theta _{ h }') -\omega (\theta _{ h })] \right\| _{ 2 } \\
& \approx \left\| \omega (\theta _{ h }')-\omega (\theta _{ h }) \right\| _{ 2 }.
\end{split}
\end{align}
\normalsize
Then, if we can design a filter $\omega$ operating on high-frequency
components so that the difference of purified Saak coefficients between
clean images $\mathbf{x}$ and adversarial images $\mathbf{x'}$ is
minimized, the difference between clean images and adversarial images
can be minimized as well. In other words, the adversarial perturbations
are removed. This will be investigated in the next subsection. 

\subsection{Saak Coefficients Filtering}

To reduce adversarial perturbations, we propose three high-frequency
Saak coefficients filtering strategies to minimize $\omega (\theta _{ h
}')-\omega (\theta _{ h })$ in this subsection.  They are dynamic-range
scaling, truncation, and clipping. Each of them will be detailed below. 

As shown in Fig.  \ref{fig:fig_01} (b), high-frequency Saak coefficients
of clean images tend to have a smaller dynamic range while those of
adversarial examples have a larger dynamic range. Thus, the first
adversarial perturbation filtering strategy is to re-scale
high-frequency Saak coefficients to match the statistics of those of
clean images. We expect adversarial perturbations to be mitigated by
enforcing the variances of high-frequency Saak coefficients of clean
images and adversarial examples to be the same. This is an empirical way
to to minimize $\left\| \omega (\theta _{ h }')-\omega (\theta _{ h })
\right\| _{ 2 }$. 

The second strategy is to truncate high-frequency Saak coefficients.
Specifically, we set the least important Saak coefficients to zeros.
Since HVS is less sensitive to high-frequency components, image
compression algorithms exploit this psycho-visual property by quantizing
them with a larger quantization step size.  In addition, high-frequency
Saak coefficients are very small for clean images, i.e. $\theta _{ h }
\approx0$ yet those of adversarial examples may become larger.  Based on
this observation, we can truncate high-frequency Saak coefficients as a
simple way to minimize $\left\| \omega (\theta _{ h }')-\omega (\theta
_{ h }) \right\| _{ 2 }$.

\begin{table*}[httttt]
\begin{center}
\scalebox{0.95}{
\begin{tabular}{l|c|c|c||c|c|c||c|c|c||c|c|c}
\toprule
Setting & \multicolumn{3}{c||}{(2-2)} & \multicolumn{3}{c||}{(2-4)} & \multicolumn{3}{c||}{(2-5)} & \multicolumn{3}{c}{(4-2)} \\ \midrule
Filtered coeff. & 40      & 42      & 44      & 640     & 660     & 680     & 2700    & 2710    & 2720    & 580     & 600     & 620     \\ \midrule
clean           & 84\%    & 84\%    & 68\%    & 96\%    & 96\%    & 91\%    & 96\%    & 95\%    & 95\%    & 96\%    & 95\%    & 91\%    \\ \midrule
FGSM            & 53\%    & 57\%    & 55\%    & 30\%    & 38\%    & 37\%    & 35\%    & 35\%    & 34\%    & 33\%    & 38\%    & 40\%    \\
BIM             & 65\%    & 69\%    & 62\%    & 49\%    & 61\%    & 61\%    & 63\%    & 63\%    & 65\%    & 54\%    & 58\%    & 64\%    \\
DF              & 77\%    & 77\%    & 65\%    & 83\%    & 83\%    & 82\%    & 86\%    & 86\%    & 86\%    & 84\%    & 83\%    & 81\%    \\
$CW_0$          & 63\%    & 61\%    & 45\%    & 56\%    & 64\%    & 66\%    & 65\%    & 62\%    & 62\%    & 58\%    & 65\%    & 64\%    \\
$CW_2$          & 84\%    & 81\%    & 66\%    & 94\%    & 92\%    & 89\%    & 93\%    & 91\%    & 90\%    & 89\%    & 93\%    & 85\%    \\
$CW_i$          & 82\%    & 80\%    & 67\%    & 90\%    & 90\%    & 85\%    & 92\%    & 92\%    & 92\%    & 87\%    & 89\%    & 85\%    \\ \midrule
All attacks     & 70.67\% & 70.83\% & 60.00\% & 67.00\% & 71.33\% & 70.00\% & \textbf{72.33\%} & 71.50\% & 71.50\% & 67.50\% & 71.00\% & 69.83\% \\
\bottomrule
\end{tabular}}
\end{center}
\caption{Comparison of the classification accuracy on the selected 
CIFAR-10 test set with four Saak transform settings under various attacks, 
where the filtering strategy is scaling the high-frequency Saak coefficients with a factor $0.25$. The obtained best result is shown in the bold font.}\label{tab:tab_01}
\end{table*}

Truncating all high-frequency components might hurt the classification
performance on clean data. The third strategy to combat adversarial
perturbations is to clip the high-frequency Saak coefficients to a
constant small value (instead of zero). 

\section{Experiments}\label{sec:experiments}

\subsection{Experimental Setup}

We conducted experiments on the CIFAR-10 and the ImageNet datasets. The
CIFAR-10 dataset \cite{krizhevsky2009learning} consists of colored
images drawn from 10 categories with a size of $32\times32$. The train
and test sets contain 50,000 and 10,000 images, respectively. The
ImageNet dataset \cite{deng2009imagenet} has 1000 classes of various
objects. It contains 1.2 million images in the training set, and 50,000
in the validation set. We apply the $224\times224$ central crop to
images from the ImageNet validation set as the input to craft
adversarial examples. 

To make a fair comparison with recent work \cite{xu2017feature,
liu2018feature}, we follow the same experiment setup of them and 
use the DenseNet \cite{huang2017densely} and the MobileNets 
\cite{howard2017mobilenets} as the target models in generating 
adversarial examples and evaluating all defense methods on CIFAR-10 
and ImageNet dataset respectively. 
We chose six popular and effective attack algorithms:
FGSM \cite{goodfellow2014explaining}, BIM \cite{kurakin2016adversarial},
DF \cite{moosavi2016deepfool}, $CW_0$, $CW_2$ and $CW_i$
\cite{carlini2017towards}, which generate adversarial examples at
different distortion constraints. The least-likely class is chosen to
generate targeted CW attacks. 
Following the setup of \cite{xu2017feature, liu2018feature},
we construct a selected set by 
taking the first 100 correctly classified samples in the test set 
as seed images to craft adversarial examples, since $CW_0$ and $CW_i$ 
attack algorithms are too expensive to run on the whole test set. 

We compare our method with recent adversarial defense methods on
mitigating adversarial effect that applied input transformations or
denoising methods as reported in \cite{xu2017feature,liu2018feature,
guo2017countering,karolina2016study}. Our method is very efficient since
it requires no adversarial training, no change on the target model and
no back-propagation to train. It can be easily implemented in Python and
we will release the code. 






\subsection{Ablation Studies}

As mentioned in \cite{kuo2018data}, the multi-stage Saak transform
provides a family of spatial-spectral representations.  The intermediate
stages in the Saak transform offer different spatial-spectral
trade-offs. We first study the effect of different hyper-parameters of
the multi-stage Saak transform. We seek to answer two questions: 1) the
spatial dimension of 3D local cuboids (LCs) in the Saak transform, and (2) the
number of stages used in the Saak transform.  We focus on four settings
of forward and inverse Saak transforms: 1) spatial dimension $2 \times 2$ in 2 stages, 
2) spatial dimension $2 \times 2$ in 4 stages, 3) spatial dimension $2 \times 2$ in 5 stages, and 4) spatial dimension $4 \times 4$ in 2 stages. For simplicity, we denote the settings
as \textit{(\{spatial dimension\}-\{stage\})}. For example, \textit{(2-5)} 
represents the setting of using local cuboid of spatial dimension $2\times2$ in 5-stage 
Saak transform. We choose the high-frequency coefficients
scaling strategy of a factor $0.25$ in all experiments and report the
CIFAR-10 classification results in Table \ref{tab:tab_01}. For each
setting, we consider 3 scenarios in terms of filtered high-frequency
Saak coefficeints. 

\begin{table*}[htttt]
\begin{center}
\scalebox{0.95}{
\begin{tabular}{l|c|c|c|c|c|c|c|c}
\toprule
Filtering & \multicolumn{4}{c|}{Truncation} & \multicolumn{4}{c}{Clipping by 0.02} \\ \midrule
Filtered coeff. & 2700    & 2710    & 2720    & 2730    & 2700    & 2710    & 2720    & 2730    \\ \midrule
clean           & 92\%    & 92\%    & 92\%    & 91\%    & 95\%    & 94\%    & 92\%    & 91\%    \\ \midrule
FGSM            & 60\%    & 63\%    & 63\%    & 62\%    & 44\%    & 47\%    & 47\%    & 44\%    \\
BIM             & 72\%    & 69\%    & 73\%    & 71\%    & 67\%    & 66\%    & 69\%    & 67\%    \\
DF              & 86\%    & 87\%    & 88\%    & 86\%    & 86\%    & 85\%    & 84\%    & 84\%    \\
$CW_0$          & 71\%    & 70\%    & 68\%    & 69\%    & 69\%    & 68\%    & 68\%    & 68\%    \\
$CW_2$          & 91\%    & 90\%    & 90\%    & 90\%    & 91\%    & 91\%    & 91\%    & 91\%    \\
$CW_i$          & 89\%    & 90\%    & 90\%    & 90\%    & 91\%    & 91\%    & 92\%    & 91\%    \\ \midrule
All attacks     & 78.17\% & 78.17\% & \textbf{78.67\%} & 78.00\% & 74.67\% & 74.67\% & 75.17\% & 74.17\% \\
\bottomrule
\end{tabular}}
\end{center}
\caption{Comparison of classification results using the
high-frequency Saak coefficient truncation and clipping strategies under
various attacks for the selected CIFAR-10 test set (with Saak setting of (2-5)). The obtained best result is shown in the bold font.} \label{tab:tab_02}
\end{table*}

Since input images of CIFAR-10 is of spatial dimension $32 \times 32$,
the spatial dimensions of the output of these settings are $8\times8$,
$2\times2$, $1\times1$, and $2\times2$, respectively. 
The corresponding spectral dimensions are 53, 853, 3413 and 785.
As shown in Table \ref{tab:tab_01}, we see better defense performance as we increase the
filter size or the stage number. The performance are comparable when the
output spatial dimensions are the same. We believe that this is related
to the receptive field size of the last stage of Saak coefficients used
for image reconstruction. The multi-stage Saak transform can incorporate
longer-distance pixel correlations to mitigate adversarial perturbations
more. For this reason, we choose the 5-stage Saak transform with local
cuboids of spatial dimension $2 \times 2$ (denoted by (2-5)) in the following experiments
as our evaluation baseline. 

\begin{table*}[httt]
\begin{center}
\scalebox{0.95}{
\begin{tabular}{l|c|c|c|c|c|c|c|c}
\toprule
Defense Methods
& FGSM & BIM & DF  & $CW_0$ &  $CW_2$  &  $CW_i$  & All attacks & clean \\ \midrule
JPEG \cite{das2017keeping,karolina2016study} (Q=90) 
& 38\% & 29\% & 67\% &  2\% & 80\% & 71\% & 47.83\% & 94\%  \\ 
Feature Distillation \cite{liu2018feature}    
& 41\% & 51\% & 79\% & 18\% & 86\% & 76\% & 58.50\% & 94\%  \\ 
Bit Depth Reduction (5-bit) \cite{xu2017feature} 
& 17\% & 13\% & 40\% &  0\% & 47\% & 19\% & 22.66\% & 93\%  \\
Bit Depth Reduction (4-bit) \cite{xu2017feature}
& 21\% & 29\% & 72\% & 10\% & 84\% & 74\% & 48.33\% & 93\%  \\
Median Smoothing (2x2) \cite{xu2017feature}
& 38\% & 56\% & 83\% & \textbf{85\%} & 83\% & 86\% & 71.83\% & 89\%  \\
Non-local Mean (11-3-4) \cite{xu2017feature} 
& 27\% & 46\% & 76\% & 11\% & 88\% & 84\% & 55.33\% & 91\%  \\ 
Cropping \cite{guo2017countering}
& 46\% & 43\% & 51\% & 15\% & 79\% & 76\% & 51.66\% & 86\%  \\
TVM      \cite{guo2017countering} 
& 41\% & 40\% & 44\% & 34\% & 75\% & 71\% & 50.83\% & 92\%  \\
Quilting \cite{guo2017countering} 
& 37\% & 42\% & 36\% & 25\% & 67\% & 70\% & 46.17\% & 90\%  \\ \midrule
Ours (4-2)
& 58\% & 70\% & 84\% & 69\% & 88\% & 88\% & 76.17\% & 90\%  \\
Ours (2-5)
& \textbf{63\%} & \textbf{73\%} & \textbf{88\%} & 68\% & \textbf{90\%} & \textbf{90\%} & \textbf{78.67\%} & 92\%  \\ \bottomrule
\end{tabular}}
\end{center}
\caption{Comparison of classification results of different defense methods 
on the selected CIFAR-10 test set.}\label{tab:tab_03}
\end{table*}

\begin{table*}[htt]
\begin{center}
\scalebox{0.95}{
\begin{tabular}{l|c|c|c|c|c|c|c|c}
\toprule
Defense Methods
& FGSM & BIM & DF  & $CW_0$ &  $CW_2$  &  $CW_i$  & All attacks & clean \\ \midrule
JPEG \cite{das2017keeping,karolina2016study} (Q=90) 
& 1\% & 0\% & 8\% & 4\% & 68\% & 32\% & 18.83\% & 70\% \\ 
Feature Distillation \cite{liu2018feature}    
& 8\% & 17\% & 55\% & 57\% & 82\% & 72\% & 48.50\% & 66\% \\ 
Bit Depth Reduction (5-bit) \cite{xu2017feature} 
& 2\% & 0\% & 21\% & 18\% & 66\% & 60\% & 27.83\% & 69\% \\
Bit Depth Reduction (4-bit) \cite{xu2017feature}
& 5\% & 4\% & 44\% & 67\% & 82\% & 79\% & 46.83\% & 68\% \\
Median Smoothing (2x2) \cite{xu2017feature}
& 22\% & 28\% & \textbf{72\%} & \textbf{85\%} & 84\% & 81\% & 62.00\% & 65\% \\
Median Smoothing (3x3) \cite{xu2017feature}
& 33\% & 41\% & 66\% & 79\% & 79\% & 76\% & 62.33\% & 62\% \\
Non-local Mean (11-3-4) \cite{xu2017feature} 
& 10\% & 25\% & 57\% & 47\% & \textbf{86\%} & 82\% & 51.17\% & 65\% \\ \midrule
Ours (4-2)
 & \textbf{47\%} & \textbf{58\%} & 65\% & 66\% & 71\% & 69\% & 62.67\% & 69\% \\
Ours (4-2) + Mean ($2\times2$)
 & 31\% & 46\% & 64\% & 68\% & 81\% & 81\% & 61.83\% & 81\% \\
Ours (4-2) + Median ($2\times2$)
 & 33\% & 45\% & 70\% & 84\% & 83\% & \textbf{83\%} & \textbf{66.33\%} & 86\% \\ \bottomrule
\end{tabular}}
\end{center}
\caption{Comparison of classification results of different defense methods 
on the selected ImageNet test set.}\label{tab:tab_04}
\end{table*}

We show the classification accuracy using the other two coefficient
filtering strategies; namely, truncation and clipping, under the same
baseline setting (2-5) in Table \ref{tab:tab_02}. By comparing Tables \ref{tab:tab_01}
and \ref{tab:tab_02}, we see that the coefficient truncation strategy
performs the best with a substantial gain over the coefficient scaling
and clipping strategies. Thus, we choose truncation strategy as the
evaluation method on CIFAR-10 and ImageNet datasets. 

\subsection{Comparison with State-of-the-Art Methods}
\begin{figure*}[ht]
\centering
\includegraphics[width=0.98\columnwidth]{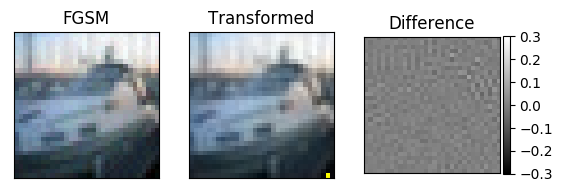} 
\includegraphics[width=0.98\columnwidth]{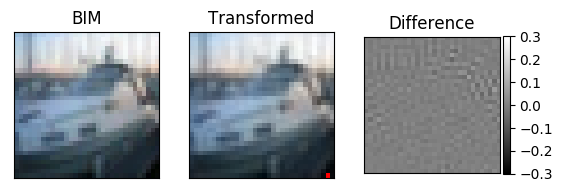} \\
\includegraphics[width=0.98\columnwidth]{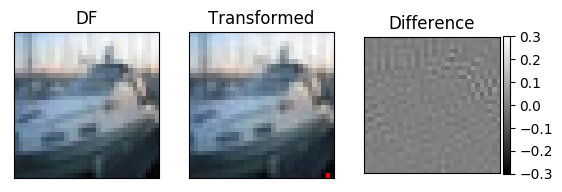}
\includegraphics[width=0.98\columnwidth]{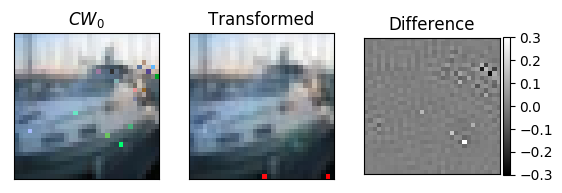} \\
\includegraphics[width=0.98\columnwidth]{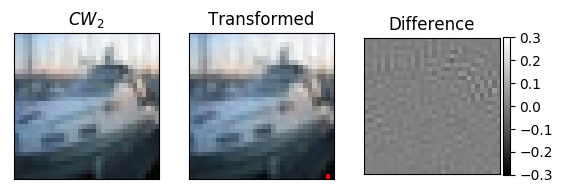} 
\includegraphics[width=0.98\columnwidth]{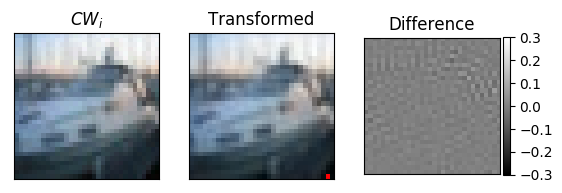} \\
\caption{Comparison of adversarial examples and Saak-transform-filtered images on six adversarial attacks with Saak transform setting (4-2) using truncation strategy. Difference image is converted into gray scale.} \label{fig:fig_02}
\end{figure*}

We compare our method with other state-of-the-art methods in combating
adversarial perturbations for the CIFAR-10 dataset in Table
\ref{tab:tab_03}. 
We see that our Saak-transform-based method
outperforms state-of-the-art methods on all attack types except for the
$CW_0$ attack by a significant margin. This may be attributed to the
fact that the $CW_0$ adversarial noise is more prominent as shown in
Fig. \ref{fig:fig_02}. It is evident that some colored patches in 
Saak-transform-filtered image are caused by $CW_0$ attack which shows that
our filtering strategy fails to remove the $CW_0$ adversarial noise well 
since it is already diffused into low-frequency channels. 
Yet, it is worthwhile to point out our proposed method achieves 90\% accuracy 
on $CW_2$ and $CW_i$ adversarial examples of CIFAR-10 dataset, which is close to
the accuracy on clean data. 
Moreover, our method is powerful in removing FGSM-based perturbations.
Furthermore, we compare different defense methods on the ImageNet dataset in
Table \ref{tab:tab_04}. Our method outperforms other defense methods 
when applied alone with the setting of (4-2). 
In addition, to verify the above conjecture on the $CW_0$
attack, we cascade the proposed method with two spatial denoising
methods, median and mean smoothing, to further improve the performance.
As shown in Table \ref{tab:tab_04}, the median smoothing method does
boost the performance on the $CW_0$ attack significantly.  Meanwhile, it
improves the classification accuracy on clean data to 86\%. These
results demonstrate that our solution is complementary to spatial smoothing techniques and, more importantly, can greatly mitigate the adversarial effect without severely hurting the classification performance on clean images unlike other methods. This is highly desirable as a computer vision system is usually unaware whether an input image is maliciously polluted or not.

\section{Conclusion}\label{sec:conclusion}

We presented a method to filter out adversarial perturbations based on
the Saak transform, a state-of-the-art spectral analysis algorithm.  
It can be used as a light-weight add-on to existing
neural networks.  The method was comprehensively evaluated in different
settings and filtering strategies. It is effective and efficient in
defending against adversarial attacks as a result of the following three
special characteristics.  It is non-differentiable. It does not modify
the target model. It requires no adversarial training and no label
information.  It was shown by experiments that the proposed method
outperforms state-of-the-art defense methods by a large margin on both
the CIFAR-10 and the ImageNet datasets while maintaining good
performance on clean images. 

\bibliographystyle{IEEEtran}
\bibliography{egbib}

\end{document}